\documentclass{article}

\usepackage{natbib}
\usepackage{enumitem}
\usepackage{caption}
\usepackage[accepted]{icml2025}

\usepackage[utf8]{inputenc}

\usepackage[T1]{fontenc}
\usepackage{times}
\usepackage{amsmath, amssymb, amsthm}
\usepackage{mathtools}
\usepackage{bbm}
\usepackage{dsfont}
\usepackage{xcolor}
\definecolor{darkpink}{RGB}{199,21,140}

\usepackage{graphicx}

\usepackage{booktabs, array}
\usepackage{multirow}

\usepackage{listings}
\usepackage{fancyvrb}
\fvset{fontsize=\small}

\definecolor{citecolor}{RGB}{0,102,204}
\definecolor{linkcolor}{rgb}{0.768, 0.054, 0.054}
\definecolor{urlcolor}{RGB}{199,21,133}

\usepackage[colorlinks,linktoc=all]{hyperref}
\usepackage[all]{hypcap}
\hypersetup{citecolor=citecolor}
\hypersetup{linkcolor=linkcolor}
\hypersetup{urlcolor=urlcolor}
\usepackage[nameinlink,capitalise]{cleveref}
\creflabelformat{equation}{#2\textup{#1}#3}  
\crefname{section}{\S}{\S\S}

\lstdefinestyle{mystyle}{
    commentstyle=\color{OliveGreen},
    numberstyle=\tiny\color{black!60},
    stringstyle=\color{BrickRed},
    basicstyle=\ttfamily\scriptsize,
    breakatwhitespace=false,
    breaklines=true,
    captionpos=b,
    keepspaces=true,
    numbers=none,
    numbersep=5pt,
    showspaces=false,
    showstringspaces=false,
    showtabs=false,
    tabsize=2
}
\newcommand{\bx}{\mathbf{x}}
\newcommand{\by}{\mathbf{y}}
\newcommand{\bz}{\mathbf{z}}

\newcommand{\calB}{{\mathcal{B}}}

\newcommand{\calD}{{\mathcal{D}}}

\newcommand{\calL}{{\mathcal{L}}}

\newcommand{\calY}{{\mathcal{Y}}}

\theoremstyle{plain}

\theoremstyle{definition}

\theoremstyle{remark}

\def\[#1\]{\begin{equation}\begin{aligned}#1\end{aligned}\end{equation}}

\newsavebox\CBox

\newcommand{\ie}{\textit{i.e.}}

\newcommand{\RETURN}{\STATE \textbf{return }}
\DeclarePairedDelimiter\ceil{\lceil}{\rceil}
\newcommand{\one}{\mathbbm{1}}

\usepackage{tcolorbox}
\newtcolorbox{prompt}[1]{colback=gray!20,colframe=gray!50!black,fonttitle=\bfseries,title=#1}

\definecolor{codeBackground}{RGB}{248,248,248} 
\definecolor{codeFrame}{RGB}{200,200,200}      
\definecolor{codeText}{RGB}{0,0,0}             
\definecolor{codeComment}{RGB}{100,100,100}    

\lstdefinestyle{myprompt}{
    backgroundcolor=\color{codeBackground},
    basicstyle=\ttfamily\small\color{codeText},
    breaklines=true,
    frame=single,
    framerule=0.5pt,
    rulecolor=\color{codeFrame},
    tabsize=4,
    keywordstyle=\bfseries,                     
    commentstyle=\itshape\color{codeComment},     
    stringstyle=\color{codeText},
    identifierstyle=\color{codeText},
    numberstyle=\color{codeText},
    showstringspaces=false,
    morecomment=[l]{\#},
}

\icmltitlerunning{FedRand: Enhancing Privacy in Federated Learning with Randomized LoRA Subparameter Updates}

\begin{document}

\twocolumn[
\icmltitle{\texorpdfstring{FedRand: Enhancing Privacy in Federated Learning \\
with Randomized LoRA Subparameter Updates}{}}


\begin{icmlauthorlist}
\icmlauthor{Sangwoo  Park}{kaist}
\icmlauthor{Seanie Lee}{kaist}
\icmlauthor{Byungjoo Kim}{kaist}
\icmlauthor{Sung Ju Hwang}{kaist,deepauto}
\end{icmlauthorlist}

\icmlaffiliation{kaist}{Graduate School of AI, KAIST}
\icmlaffiliation{deepauto}{DeepAuto.ai}

\icmlcorrespondingauthor{Sangwoo Park}{swgger@kaist.ac.kr}
\icmlcorrespondingauthor{Seanie Lee}{lsnfamily02@kaist.ac.kr}

\icmlkeywords{Machine Learning, ICML}

\vskip 0.3in
]



\printAffiliationsAndNotice{}  

\begin{abstract}
Federated Learning (FL) is a widely used framework for training models in a decentralized manner, ensuring that the central server does not have direct access to data from local clients. However, this approach may still fail to fully preserve data privacy, as models from local clients are exposed to the central server during the aggregation process. This issue becomes even more critical when training vision-language models (VLMs) with FL, as VLMs can easily memorize training data instances, making them vulnerable to membership inference attacks (MIAs). To address this challenge, we propose the \emph{FedRand} framework, which avoids disclosing the full set of client parameters. In this framework, each client randomly selects subparameters of Low-Rank Adaptation (LoRA) from the server and keeps the remaining counterparts of the LoRA weights as private parameters. After training both parameters on the client's private dataset, only the non-private client parameters are sent back to the server for aggregation. This approach mitigates the risk of exposing client-side VLM parameters, thereby enhancing data privacy. We empirically validate that FedRand improves robustness against MIAs compared to relevant baselines while achieving accuracy comparable to methods that communicate full LoRA parameters across several benchmark datasets.
\end{abstract}
\section{Introduction}
Vision-language models (VLMs) \citep{flamingo, minigpt, llava} have demonstrated remarkable performance in various multi-modal tasks, such as visual question answering \citep{instructblip, llava} and image captioning \citep{blip}. However, deploying VLMs in real-world scenarios raises significant concerns about data privacy. These models can easily memorize training data \citep{carlini2021extracting, carlini2023quantifying}, including sensitive information such as private photographs or medical diagnosis records. Adversarial attackers can exploit this vulnerability to perform a membership inference attack~\citep{mia}, which aims to detect whether a specific data instance is part of the training dataset.

Federated learning~\citep[FL;][]{fed-avg} is a distributed learning framework in which each local client receives global parameters from a central server, trains a local model on its private dataset, and periodically sends the local model back to the server for aggregation. It offers a potential solution to address privacy concerns, as the central server cannot directly access the private dataset. However, naively transmitting local model parameters back to the central server remains vulnerable to membership inference attacks, as attackers can potentially reconstruct the local client model by intercepting its parameters during the aggregation stage. This issue is particularly critical when fine-tuning vision-language models (VLMs), as their large capacity to memorize private training data amplifies the privacy risks.

\begin{figure*}[t]
\centering
    \includegraphics[width=1.0\textwidth]{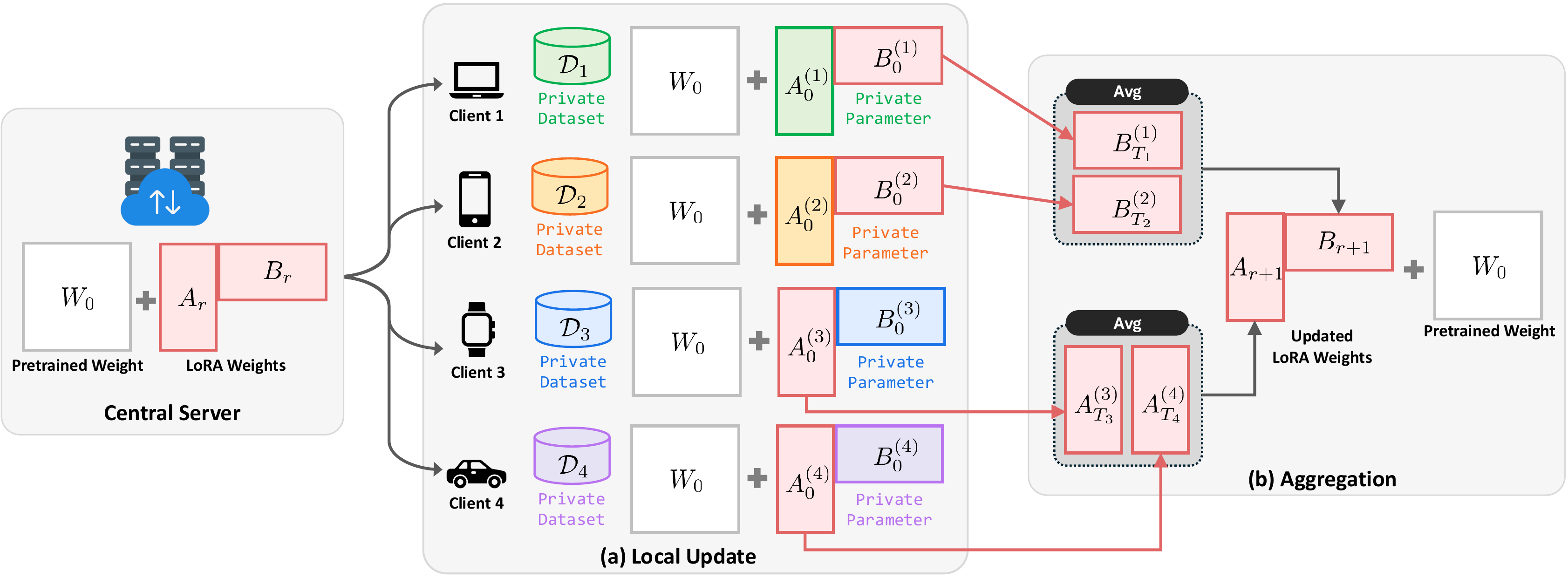}
    \vspace{-0.1in}
    \caption{\textbf{(a)}. At each round $r$, each local client selects a LoRA weight either $A_r$ or $B_r$ for initialization from the server and initializes the other counterparts of LoRA weights using the previous round's client model parameters as private parameters. \textbf{(b)}. After updating both parameters, only the non-private parameters are sent back to the server and aggregated to update the LoRA weights of the central server.}
    \label{fig:concept-fig}
    \vspace{-0.1in}
\end{figure*}

To address the privacy issue, we propose a simple yet privacy-enhanced federated learning (FL) framework, dubbed \emph{FedRand}. In this framework, clients randomly select a subset of parameters provided by the server and keep the remaining parameters as client-specific private ones. After updating both the selected parameters and their client-specific private parameters, only the non-private parameters are transmitted back to the server for the model update.

Specifically, we first apply Low-Rank Adaptation~\citep[LoRA;][]{lora} matrices $A$ and $B$ to the pre-trained weight $W_0$ of a VLM. The pre-trained weight $W_0$ is fixed and shared across all clients and the server. At each round of updates, each local client model receives the LoRA weights $A$ and $B$ from the server. Each client then randomly selects either $A$ or $B$ and initializes the counterpart of the LoRA weights using the parameters from the previous round as client-specific private ones(\Cref{fig:concept-fig}\textcolor{linkcolor}{a}). After updating both parameters on the client’s private training dataset, the client-specific parameters remain hidden, and only the remaining parameters are sent back to the server. Finally, the parameters $A$ and $B$ from the clients are averaged to form the new LoRA weights of the server model (\Cref{fig:concept-fig}\textcolor{linkcolor}{b}). Since the client-specific private parameters are kept hidden, adversarial attackers cannot fully reconstruct the client model parameters by intercepting the parameters transmitted to the server. This design makes FedRand more robust against membership inference attacks. Furthermore, sending only non-private parameters significantly reduces the communication cost between the server and clients compared to the model that communicates all LoRA weights between the server and clients.

\begin{figure}[t]
\centering
    \includegraphics[width=0.35\textwidth]{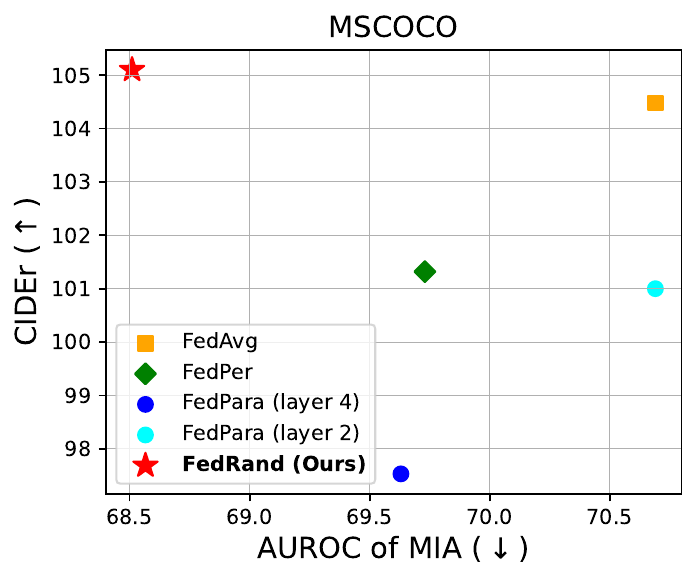}
    \vspace{-0.1in}
    \caption{Trade-off between task performance (CIDEr) and vulnerability to membership inference attacks (AUROC of MIA) on MSCOCO dataset.}
    \label{fig:trade-off}
\vspace{-0.1in}
\end{figure}

We empirically validate our proposed FedRand on visual question answering and image captioning tasks using the ScienceQA~\citep{scienceqa}, MSCOCO~\citep{mscoco}, and NoCaps~\citep{nocaps} datasets. Experimental results demonstrate that FedRand significantly improves the trade-off between accuracy and robustness against membership inference attacks (\Cref{fig:trade-off}) while reducing communication costs between the server and clients compared to other relevant baselines.

Our contributions and findings are summarized below:
\begin{itemize}
[itemsep=1mm,parsep=1pt,topsep=0pt,leftmargin=*]

\item We show that even fine-tuning VLMs with FL remains vulnerable to membership inference attacks due to the exposure of client model parameters, posing significant privacy concerns.

\item To address these privacy concerns, we propose FedRand. First, a client randomly selects subparameters of LoRA weights from the server and updates both the selected parameters and client-specific private parameters. Only the non-private parameters are sent back to the server, preventing the exposure of the full local model parameters.

\item We experimentally demonstrate that FedRand enhances robustness against membership inference attacks while achieving performance comparable to models that communicate full LoRA weights between the server and clients.
\end{itemize}
\section{Related Work}
\paragraph{Federated learning.}
Federated Learning (FL) is a decentralized machine learning approach that allows multiple clients to collaboratively train a shared model without sharing their private data, thereby preserving privacy and security. FedAvg~\citep{fed-avg}, one of the most widely used algorithms in FL, updates a global model by averaging the model parameters trained on each client's private dataset. While many variants of FedAvg have been proposed~\citep{li2020federated, yu2020federated, acar2021federated, fedit}, they remain vulnerable to membership inference attacks because clients' parameters are exposed to the server. In another line of work, methods like FedPer~\citep{fedper} and FedPara~\citep{fedpara} distinguish client-specific private parameters from global parameters shared between the server and clients to reduce communication costs. However, although these methods avoid exposing client parameters to the server, they fail to strike a balance between accuracy and robustness against membership inference attacks.

\paragraph{Membership inference attack.}
Although FL avoids sharing private data between clients and a server by training client models locally and aggregating only the parameters of the client models at the server, clients are still vulnerable to the leakage of privacy-sensitive information. This can occur through membership inference attacks~\citep{mia}, where an attacker detects whether a specific data instance is included in a private client's dataset. While both the central server and clients can potentially deduce private details from shared information such as model parameters, the majority of works~\citep{hitaj2017deep, melis2019exploiting} focus on client-based membership inference attacks under the strong assumption of a secure server. However, server-based membership inference attacks pose a significant threat, particularly due to the memorization capacities of VLMs. \citet{dejavu} have demonstrated this vulnerability through \textit{k}-nearest neighbor retrieval tests on open-source image datasets, showing that VLMs are prone to retaining training data. Moreover, \citet{li2024membership} utilize average top-k Rényi entropy of VLMs' output probabilities to distinguish training data from other data, highlighting the vulnerability of VLMs to membership inference attacks. This suggests that malicious use of client models on the server side could lead to data leakage through the memorization of training data by the client models. To address this issue, we propose FedRand, which prevents the exposure of client models to the server and thus enhances robustness against server-based membership inference attacks.

\section{Method}
\subsection{Preliminaries}
Let $p_\theta: \mathcal{X}\times \mathcal{Z}\to\mathcal{Y}$ be a vision language model (VLM) with its parameter $\theta$, which takes as input a sequence of tokens $\bx\in\mathcal{X}$ and an image $\bz\in\mathcal{Z}$, and outputs another sequence of tokens $\by\in\mathcal{Y}$ as a response to the input. Here, $\mathcal{X}$ is the set of all possible input sequences, $\mathcal{Z}$ is the set of all possible images, and $\calY$ is the set of all possible output sequences. In the FL framework, each client $k\in [K]\coloneqq\{1,\ldots, K\}$ has access only to its local training dataset $\mathcal{D}_k=\{(\bx^{(k)}_i, \bz^{(k)}_i,  \by^{(k)}_i) \}_{i=1}^{n_k}$, where $\calD_k \cap \calD_{k^\prime}=\emptyset$ for all $k,k^\prime \in [K]$ with $k\neq k^\prime$. Furthermore, the central server does not have direct access to any of the local datasets. 
For each round of update $r\in [R]$, a subset of client indices $S_r\subset [K]$ is randomly chosen with $\lvert S_r\rvert = K^\prime$. Then each client $k\in S_r$ receives the parameter $\theta_r$ from the central server and trains its local model $p_{\theta^{(k)}}$ on the dataset $\calD_k$ as follows:
\begin{equation}
\begin{gathered}
\theta^{(k)}_{r, t+1} = \theta^{(k)}_{r,t} - \eta \nabla_\theta \mathcal{L}(\theta^{(k)}_{r,t}; \calD_{k}) \\
\calL(\theta^{(k)}_{r,t};\calD_k)= -\frac{1}{n_k}\sum_{(\bx,\bz,\by)\in\calD_k} \log p_{\theta_{r,t}^{(k)}}(\by\mid \bx, \bz),
\end{gathered}
\label{eq:obj}
\end{equation}
for $t=0, \ldots, T_k-1$, where $\eta>0$ is a learning rate and $\theta^{(k)}_{0,0}$ is initialized with $\theta_r$. Since fully fine-tuning the VLM is computationally expensive, we apply Low Rank Adaptation ~\citep[LoRA;][]{lora} for fine-tuning the weight matrix of the VLM at the $l$-th layer as:
\begin{equation}
    W^{(k,l)}_{r,t} = W^{(l)}_0 + A^{(k,l)}_{r,t}B^{(k,l)}_{r,t},
\end{equation}
where $W^{(l)}_0$ is the frozen pre-trained weight matrix of the VLM, and $A^{(k,l)}_{r,t}$ and $B^{(k,l)}_{r,t}$ are low-rank matrices, \ie, $\text{rank}(A^{(k,l)}_{r,t}B^{(k,l)}_{r,t}) \ll \text{rank}(W^{(l)}_0)$. 
With a slight abuse of notation of $\theta^{(k)}_{r,t}$, we denote the parameter $\theta^{(k)}_{r,t}=\{(W^{(l)}_0, A^{(k,l)}_{r,t}, B^{(k,l)}_{r,t}) \}_{l=1}^L$ as the set of the initial pre-trained weight matrices and LoRA weight matrices for the client $k$ at step $t$ in round $r$. After the local client update, following the FedAvg~\citep{fed-avg} and FedIT~\citep{fedit}, we aggregate the parameters of the local client models and update the server parameter $\theta_r=\{(W^{(l)}_0, A^{(l)}_r, B^{(l)}_r)\}_{l=1}^L$ to $\theta_{r+1}$ as follows:
\begin{align}
    A^{(l)}_{r+1} &= \left(\sum_{k\in S_r}\frac{n_k}{m_r}A^{(k, l)}_{r,T_k}\right), B^{(l)}_{r+1} &=\left(\sum_{k\in S_r}\frac{n_k}{m_r}B^{(k, l)}_{r,T_k}\right)
\end{align}    
where $m_r= \sum_{k\in S_r}n_k$ and $n_k=\lvert \mathcal{D}_k \rvert$. At the next round $r+1$, the central server model $p_{\theta_{r+1}}$ uses its updated weight matrix,
\begin{equation}
    W^{(l)}_{r+1} = W^{(l)}_0 + A^{(l)}_{r+1}B^{(l)}_{r+1}
\end{equation}
for each layer $l\in [L]$.

\begin{figure}[t]
\vspace{-0.15in}
\centering
\begin{minipage}{0.4865\textwidth}
\begin{algorithm}[H]
   \caption{\small FedRand}
   \begin{algorithmic}[1]
    \STATE \textbf{Input}: VLM $p_\theta$ with pre-trained weights $\theta=\{W^{(l)}_0\}_{l=1}^L$, learning rate $\eta$, total round $R$, number of clients $K$, number of clients participating for update $K^\prime$, probability $\rho$ of choosing $A$, and batch size $b$. 
    \STATE Randomly initialize LoRA weights $\{(A^{(l)}_0, B^{(l)}_0)\}_{l=1}^L$.
    \FOR{$r=0,\ldots,R-1$}
    \STATE $m_r \leftarrow 0$, $\theta_r \leftarrow \{(W^{(l)}_0, A^{(l)}_r, B^{(l)}_r) \}_{l=1}^L$
    \STATE Choose client indices $S_r$ from $[K]$ s.t. $\lvert S_r\rvert=K^\prime$.
    \FOR{each $k$ in $S_r$}
    \STATE $(\theta^{(k)}, a_k, n_k) \leftarrow \text{client\_update}(k, \theta_r, E, \rho, \eta, b, r)$
    \STATE $m_r \leftarrow m_r + n_k$
    \ENDFOR
    \STATE $\alpha \leftarrow \sum\limits_{k\in S_r}  \frac{n_k}{m_r} \cdot\one_{\{a_k=1\}}$,
    $\beta \leftarrow \sum\limits_{k\in S_r}\frac{n_k}{m_r}\cdot \one_{\{a_k\neq1\}}$
    \FOR{$l=1,\ldots, L$}
    \IF{$\alpha >0$}
    \STATE $A^{(l)}_{r+1} \leftarrow \sum_{k\in S_r, a_k=1} \frac{n_k}{\alpha m_r} A^{(k,l)}_{r,T_k}$
    \ELSE
    \STATE $A^{(l)}_{r+1} \leftarrow A^{(l)}_{r}$
    \ENDIF
    \IF{$\beta >0$}
    \STATE $B^{(l)}_{r+1} \leftarrow \sum_{k\in S_r, a_k\neq1} \frac{n_k}{\beta m_r} B^{(k,l)}_{r,T_k}$
    \ELSE
    \STATE $B^{(l)}_{r+1} \leftarrow B^{(l)}_{r}$
    \ENDIF
    \ENDFOR
    \ENDFOR
    \STATE $\theta_* \leftarrow \{(W^{(l)}_0, A^{(l)}_R, B^{(l)}_R) \}_{l=1}^L$
    \RETURN $p_{\theta_*}$
   \end{algorithmic}
\label{algo-1}
\end{algorithm}
\end{minipage}
\vspace{-0.2in}
\end{figure}

\begin{figure}[t]
\vspace{-0.15in}
\centering
\begin{minipage}{0.48\textwidth}
\begin{algorithm}[H]
   \caption{\small client\_update($k$, $\theta$, $E$, $\rho$, $\eta$, $b$, $r$)}
   \begin{algorithmic}[2]
    \STATE \textbf{Input}: Client index $k$, server parameter $\theta_r = \{(W^{(l)}_0, A^{(l)}_r, B^{(l)}_r)\}_{l=1}^L$, train epochs $E$,  probability $\rho$ of choosing $A^{(l)}$, learning rate $\eta$, batch size $b$, and current round $r$.
    \STATE $T_k\leftarrow \ceil*{\lvert \calD_k \rvert / b} \cdot E$
    \STATE $u_k\leftarrow \text{Uniform}(0,1)$, $a_k\leftarrow \one_{\{u_k <\rho \} }$
    \IF{$r=0$}
    \STATE $\{A^{(k,l)}_{0,0}\}_{l=1}^L \leftarrow \{\texttt{rand\_init}(A^{(l)}_{r}) \}_{l=1}^L $
    \STATE $\{B^{(k,l)}_{0,0}\}_{l=1}^L \leftarrow \{\texttt{zero\_init}(B^{(l)}_{r}) \}_{l=1}^L $
    \ELSE
        \IF{$a_k=1$}
        \STATE $\{A^{(k,l)}_{r,0}\}_{l=1}^L\leftarrow \{A^{(l)}_r\}_{l=1}^L$
        \STATE $\{B^{(k,l)}_{r,0}\}_{l=1}^L\leftarrow \{B^{(k,l)}_{r-1, T_k}\}_{l=1}^L$
        \ELSE
        \STATE $\{A^{(k,l)}_{r,0}\}_{l=1}^L\leftarrow \{A^{(k,l)}_{r-1, T_k}\}_{l=1}^L$
        \STATE $\{B^{(k,l)}_{r,0}\}_{l=1}^L\leftarrow \{B^{(l)}_r\}_{l=1}^L$
        \ENDIF
    \ENDIF
    \FOR{$t=0,\ldots, T_k-1$}
    \STATE Sample a mini-batch $\calB$ from the client dataset  $\calD_k$.
    \STATE $\theta^{(k)}_{r,t}\leftarrow\{(W^{(l)}_0, A^{(k,l)}_{r,t}, B^{(k,l)}_{r, t})\}_{l=1}^L$ 
    \STATE $\calL(\theta^{(k)}_{t}; \mathcal{B}) \leftarrow -\frac{1}{\lvert\calB \rvert}\sum_{(\bx,\bz, \by)\in \calB}\log p_{\theta^{(k)}_{t}}(\by\mid \bx, \bz)$
    \STATE $\theta^{(k)}_{r,t+1}\leftarrow \theta^{(k)}_{r,t} -\eta \nabla_{\theta^{(k)}_{r,t}} \calL(\theta^{(k)}_{r,t} ;\calB)$
    \ENDFOR
   \STATE Cache $\{(A^{(k,l)}_{r, T_k}, B^{(k,l)}_{r, T_k})\}$
   \IF{$a_k =1$ }
   \RETURN $\left(\{A^{(k,l)}_{T_k}\}_{l=1}^L, a_k, \lvert \calD_k\rvert\right)$ 
   \ELSE
   \RETURN $\left(\{B^{(k,l)}_{T_k}\}_{l=1}^L, a_k, \lvert \calD_k\rvert\right)$ 
   \ENDIF
   \end{algorithmic}
\label{algo-client}
\end{algorithm}
\end{minipage}
\vspace{-0.2in}
\end{figure}

\subsection{Privacy Enhanced FL: FedRand}
However, aggregating the parameters of client models at the central server poses a serious privacy issue. An adversarial attacker can fully reconstruct the local model by hijacking the LoRA parameters. Since VLMs easily memorize training data~\citep{carlini2021extracting, carlini2023quantifying, dejavu}, the attacker can detect whether a particular training data instance is included in the local client's training dataset $\calD_k$ using a membership inference attack~\citep{mia, li2024membership}.

To address the issue of exposing the full parameters of local client models to an attacker, we propose \emph{FedRand}, a method in which, during each update round, each client randomly selects either $\{A^{(l)}_r\}_{l=1}^L$ or $\{B^{(l)}_r\}_{l=1}^L$ LoRA weights from the server as initialization, while the remaining components are initialized using the previous round's client model parameters $\theta^{(k)}_{r-1,T_k}=\{(W_0^{(l)}, A^{(k,l)}_{r-1, T_k}, B^{(k,l)}_{r-1, T_k}) \}_{l=1}^L$ as private parameters. Only the selected parameters are sent back to the server after updating the client model, whereas the client-specific private LoRA weights remain hidden. This randomized LoRA subparameter update prevents the attacker from fully recovering the parameters of the local client model, thereby enhancing robustness against membership inference attacks. Furthermore, our proposed method, FedRand, helps save communication costs by reducing the number of parameters sent from clients to the server compared to the FedAvg method.

Specifically, at each round $r\in [R]$, each client $k\in S_r$ first samples $a_k$ with a probability $\rho$ of choosing $\{A^{(l)}_r \}_{l=1}^L$ as follows:
\begin{equation}
    u^{(k)} \sim \text{Uniform}(0,1), \quad a_k = \one_{\{u^{(k)} < \rho\}},
\end{equation}
where $\one$ is an indicator function. The binary variable $a_k\in \{0,1\}$ indicates whether $A^{(l)}_r$ is selected.
If $a_k = 1$, we initialize $A^{(k,l)}_{r,0}$ with $A^{(l)}_r$ from the server and randomly initialize its counterpart, $B^{(k,l)}_{r,0}$, with the client parameter $B^{(k,l)}_{r-1, T_k}$ from the previous round $r-1$. Otherwise, we reverse the procedure as follows:
\begin{gather}
    A^{(k,l)}_{r, 0} =
    \begin{cases}
        A^{(l)}_r, & \text{if } a_k =1, \\
        A^{(l)}_{r-1, T_k}, & \text{otherwise},
    \end{cases} \\
    B^{(k,l)}_{r,0} =
    \begin{cases}
        B^{(k,l)}_{r-1, T_k}, & \text{if } a_k = 1, \\
        B^{(l)}_r, & \text{otherwise}.
    \end{cases}
\end{gather}
for all layers $l \in [L]$. Note that $A_{0,0}^{(k,l)}$ is randomly initialized and $B_{0,0}^{(k,l)}$ is initialized as a zero matrix, regardless of the choice of $a_k$. Then, we update the local client model, initialized with $\theta^{(k)}_0=\{(W^{(l)}_0, A^{(k,l)}_0, B^{(k,l)}_0) \}_{l=1}^L$, as described in~\Cref{eq:obj}, for $T_k$ steps, yielding $\theta^{(k)}_{T_k}=\{(W^{(l)}_0, A^{(k,l)}_{T_k}, B^{(k,l)}_{T_k}) \}_{l=1}^L$. After the local update, only the selected LoRA parameters are sent back to the central server and the parameter of the central server model is updated to $\theta_{r+1}=\{(W^{(l)}_0, A^{(l)}_r, B^{(l)}_r) \}_{l=1}^L$ as follows:
\begin{gather}
\alpha = \sum_{k\in S_r}\frac{n_k}{m_r} \cdot \one_{\{a_k=1\}}, \quad \beta = \sum_{k\in S_r}\frac{n_k}{m_r} \cdot \one_{\{a_k\neq1\}} \label{eq:normalization} \\
    A^{(l)}_{r+1} = \begin{cases}
        \sum_{k\in S_r, a_k=1} \frac{n_k}{\alpha m_r} A^{(k,l)}_{r, T_k}, &\text{if } \alpha >0\\
        A^{(l)}_r, &\text{otherwise},
     \end{cases} \label{eq:agg-A} \\
B^{(l)}_{r+1} = \begin{cases}
        \sum_{k\in S_r, a_k\neq1} \frac{n_k}{\beta m_r} B^{(k,l)}_{r, T_k}, &\text{if } \beta >0\\
        B^{(l)}_r, &\text{otherwise},
     \end{cases} \label{eq:agg-B}
\end{gather}
where $m_r = \sum_{k\in S_r}n_k$ and $n_k = \lvert \calD_k\rvert$. The parameters  $\{A^{(k,l)}_{r, T_k}\}_{l=1}^L$ are aggregated from the clients whose $a_k=1$, while $\{B^{(k,l)}_{r, T_k}\}_{l=1}^L$ are aggregated from the clients whose $a_k\neq1$. If none of the clients choose the server parameters $\{A^{(l)}_r\}_{l=1}^L$, the parameters are not updated and remain the same for $\{A^{(l)}_{r+1}\}_{l=1}^L$. The same rule applies to the update of $\{B^{(l)}_r\}_{l=1}^L$.  Note that we need normalization factors $\alpha$ and $\beta$ to  ensure that the summation of the coefficients in~\Cref{eq:agg-A} and~\Cref{eq:agg-B} equals one, respectively. Otherwise, the summation of coefficients would not equal to one, since some of the weight matrices from the clients are not sent back to the server. After $R$ rounds of updates, we use $\theta_*=\{(W_0^{(l)}, A^{(l)}_R, B^{(l)}_R) \}_{l=1}^L$ as the parameters of the final server model $p_{\theta_*}$. We outline our method in~\Cref{algo-1} and~\Cref{algo-client}.

\section{Experiments}
\subsection{Setup}
\paragraph{Dataset.}
To evaluate both the effectiveness and privacy robustness of FedRand, we conduct two experiments:
(a) accuracy evaluation on visual question answering (VQA) and image captioning tasks, and
(b) a membership inference attack using models trained in experiment (a). For the VQA task, we use the ScienceQA~\citep{scienceqa} dataset, while for the image captioning task, we use MSCOCO~\citep{mscoco}. To assess out-of-distribution (OOD) generalization and robustness against membership inference attacks, we employ the NoCaps~\citep{nocaps} dataset. For the non-IID scenarios, we use the Dirichlet distribution to randomly split each dataset, where ScienceQA is divided based on topics, while MSCOCO is partitioned according to object classes in images. We set the Dirichlet parameter to $0.5$ as suggested by FedML~\citep{FedML}. Detailed descriptions of each dataset can be found in \Cref{sec:appendix-a-1}.

\paragraph{Evaluation metrics.}
For ScienceQA dataset, we measure the exact match between ground truth answers and  model predictions as an accuracy. 
For MSCOCO and NoCaps datasets, BLEU~\citep{bleu}, ROUGE~\citep{rouge}, and CIDEr~\citep{cider} score are utilized to evaluate the quality of the responses. Lastly, we use the MaxRényi-K\%~\citep{li2024membership} metric as a score for binary classification between member and non-member data, defined as follows: 
\begin{equation}
\begin{gathered}
    \text{MaxRény-K\%}(X) \\= \frac{1}{\lvert \text{Max-K\%}(X) \rvert} \sum_{i\in \text{Max-K\%}(X)}H_\alpha (p_\theta(\cdot \mid x_{1:i})),
\end{gathered}
\label{eq:max-renyi}
\end{equation}
where $X=(x_1, \ldots, x_T)$ is an input token sequence, $p_\theta(\cdot\mid x_{1:i})$ denotes the next-token distribution after the $i$-th token, and $\text{Max-K\%}(X)$ is the set of token positions in $X$ with the highest $K\%$ Rény entropy $H_\alpha$. With this score, we compute the AUROC score to measure the robustness against membership inference attacks.   
Note that MaxRény-0\% is the maximum Rényi entropy among all positions from 1 to $T-1$, \ie, $\max_{i\in [T-1]} H_\alpha(p_\theta(\cdot\mid x_{1:i}))$.

\begin{table*}[t]
    \centering
    \footnotesize
    \caption{We train each method on the VQA, ScienceQA, and MSCOCO datasets and report its performance on the server, as well as the average performance of the clients. The best results are \textbf{bolded}, and the second-best ones are \underline{underlined}.}
    \label{tab:vl-acc}
    
    \resizebox{0.9\textwidth}{!}{\begin{tabular}{lccccccc}
        \toprule
        \textit{Server}
        
        & \textbf{ScienceQA}
        & \multicolumn{5}{c}{\textbf{MSCOCO}} \\

        \midrule

        \textbf{Method}
        & Acc
        & BLEU-1 & BLEU-2 & BLEU-3 & BLEU-4 & ROUGE & CIDEr \\

        \midrule

        FedAvg (oracle)  & \textbf{81.50} {\scriptsize (0.53)} & \textbf{75.49} {\scriptsize (0.44)} & \underline{58.53} {\scriptsize (0.37)} & \underline{43.43} {\scriptsize (0.32)} & \underline{31.80} {\scriptsize (0.17)} & \textbf{55.29} {\scriptsize (0.22)} & \textbf{111.08} {\scriptsize (0.76)} \\

        FedPer (2 layer)  & 42.11 & 74.53 & 57.11 & 41.97 & 30.12 & 54.13 & 106.60 \\

        FedPer (4 layer)  & 44.59 & 74.43 & 57.31 & 42.14 & 30.22 & 54.17 & 107.44 \\

        FedPara  & 64.78 & 73.73 & 56.94 & 41.36 & 29.91 & 53.75 & 106.96 \\
        \midrule
        \textbf{FedRand (Ours)}  & \underline{80.12} {\scriptsize (0.42)} & \underline{75.37} {\scriptsize (0.35)} & \textbf{58.66} {\scriptsize (0.38)} & \textbf{43.63} {\scriptsize (0.23)} & \textbf{31.89} {\scriptsize (0.25)} & \underline{55.15} {\scriptsize (0.19)} & \underline{110.27} {\scriptsize (0.54)} \\
        \bottomrule
        \toprule

        \textit{Client}
        & \textbf{ScienceQA}
        & \multicolumn{5}{c}{\textbf{MSCOCO}} \\

        \midrule

        \textbf{Method}
        & Acc
        & BLEU-1 & BLEU-2 & BLEU-3 & BLEU-4 & ROUGE & CIDEr \\


        \midrule

        FedAvg (oracle)  & \textbf{79.90} {\scriptsize (1.26)} & \underline{73.86} {\scriptsize (0.56)} & \underline{56.62} {\scriptsize (0.57)} & \underline{41.43} {\scriptsize (0.54)} & \underline{29.76} {\scriptsize (0.51)} & \textbf{54.04} {\scriptsize (0.32)} & \underline{104.48} {\scriptsize (1.21)} \\

        FedPer (2 layer)  & 56.93 {\scriptsize (5.40)} & 71.82 {\scriptsize (1.52)} & 54.20 {\scriptsize (1.78)} & 39.10 {\scriptsize (1.61)} & 27.67 {\scriptsize (1.35)} & 52.45 {\scriptsize (0.93)} & 101.00 {\scriptsize (3.63)} \\

        FedPer (4 layer)  & 58.94 {\scriptsize (5.73)} & 72.52 {\scriptsize (1.39)} & 54.99 {\scriptsize (1.61)} & 39.84 {\scriptsize (1.51)} & 28.34 {\scriptsize (1.26)} & 52.96 {\scriptsize (0.72)} & 101.32 {\scriptsize (2.86)} \\

        FedPara  & 58.57 {\scriptsize (5.20)} & 71.36 {\scriptsize (1.60)} & 53.51 {\scriptsize (2.18)} & 38.25 {\scriptsize (2.13)} & 26.86 {\scriptsize (1.69)} & 52.90 {\scriptsize (1.23)} & 97.53 {\scriptsize (5.03)} \\

        \midrule
        \textbf{FedRand (Ours)}  & \underline{76.01} {\scriptsize (1.15)} & \textbf{73.90}  {\scriptsize (0.89)} & \textbf{56.76} {\scriptsize (0.97)} & \textbf{41.72} {\scriptsize (0.94)} & \textbf{29.94} {\scriptsize (0.63)} & \underline{53.64} {\scriptsize (0.69)} & \textbf{105.10} {\scriptsize (1.30)} \\
        \bottomrule

    \end{tabular}
}
\vspace{-0.1in}
\end{table*}

\paragraph{Implementation details.} We use a pre-trained model trained with the TinyLLava~\citep{tinyllava} framework, which consists of an image encoder, CLIP~\citep{clip}, an instruction-tuned language model, OpenELM~\citep{openelm}, with 450M parameters, and a linear transformation layer that maps the output of CLIP to the word embedding space of OpenELM. We fine-tune only the language model using LoRA with a rank of 8, while keeping the rest of the model frozen. For each round of FL updates, we fine-tune a client model using the AdamW~\citep{adamw} optimizer for one epoch, with a learning rate of $3\cdot 10^{-4}$, weight decay of $10^{-6}$, a batch size of 8, and $\rho=0.5$. We set the total number of clients $K$ to 12 and sample $30\%$ of clients at each round during FL (\ie, $K^\prime=4$). The total number of FL update rounds is set to 30.

\paragraph{Baselines.}
We compare our proposed method, FedRand, against the following relevant baselines.
\begin{enumerate}
[itemsep=1mm,parsep=1pt,topsep=0pt,leftmargin=*]
    \item \textbf{FedAvg}~\citep{fed-avg} trains local clients using the full LoRA weights provided by a central server and averages the updated full LoRA weights from clients to update the server model's parameters.
    
    \item \textbf{FedPer}~\citep{fedper} communicates the LoRA weights of certain top layers between the server and clients while keeping the remaining LoRA weights as client-specific private parameters. We share the top 2 or 4 layers of LoRA weights across clients as global parameters. The other layers of LoRA weights are kept hidden as client-specific private parameters and are never shared. Since LoRA parameters of certain layers remain entirely private in FedPer, the LoRA A and B matrices of these non-shared layers were initialized using the aggregation results from the first round to ensure training stability.
    
    \item \textbf{FedPara}~\citep{fedpara} parameterizes private LoRA weights for each client and global LoRA weights shared across the server and clients. Each client performs elementwise multiplication between its private LoRA weights and the global ones, then adds the result to the initial pre-trained weights. The global parameters are aggregated from the clients and averaged to serve as the parameters of the server model.

\end{enumerate}
FedAvg serves as the oracle method for accuracy evaluation experiments, as it always communicates the full LoRA weights between the server and clients. The other two baselines are selected because they share the concept of partial parameter sharing with our method, enabling a comparative analysis of different strategies. The details of the implementation for FedPer and FedPara are provided in \Cref{sec:appendix-a-4}.

\subsection{Experimental Results}
\paragraph{Main results.} \Cref{tab:vl-acc} presents the performance of FedRand and other baselines on the  ScienceQA and MSCOCO datasets. The upper table reports the statistics of the server-side aggregated global model, while the lower table summarizes the average statistics of individual client models. Given the dynamic client participation in FL, we conducted three runs with different random seeds for the top two performing methods: FedAvg and FedRand. On both the server and client sides, the results indicate that FedRand achieves comparable performance to FedAvg — an oracle method that communicates full LoRA parameters between the server and clients in every round without considering membership inference attacks. This highlights the effectiveness of our proposed method, FedRand, while reducing communication costs between the server and clients by sharing only a subset of client parameters in each round.

\begin{table}[t]
\vspace{-0.1in}
    \centering
    \caption{We evaluate each method trained on the MSCOCO dataset to measure OOD generalization on the NoCaps dataset.}
    \label{tab:vl-ood}
    \resizebox{0.47\textwidth}{!}{\begin{tabular}{lcccccc}
        \toprule
        \textit{Server}
        & \multicolumn{6}{c}{\textbf{NoCaps}} \\

        \midrule

        \textbf{Method}
        & BLEU-1 & BLEU-2 & BLEU-3 & BLEU-4 & ROUGE & CIDEr \\


        \midrule

        FedAvg (oracle) & \underline{78.74} & \textbf{62.24} & \underline{46.38} & \underline{33.49} & \textbf{54.66} & \textbf{79.82} \\


        FedPer (2 layer) & 78.10 & 61.30 & 45.30 & 32.20 & 53.20 & 78.10 \\

        FedPer (4 layer) & 78.40 & 61.80 & 45.80 & 32.80 & 54.30 & 78.50 \\

        FedPara & 77.10 & 59.90 & 43.90 & 31.10 & 53.40 & 77.20 \\
        \midrule
        \textbf{FedRand (Ours)} & \textbf{78.81} & \underline{62.23} & \textbf{46.42} & \textbf{33.61} & \underline{54.57} & \underline{79.23} \\
        
        \bottomrule
    \end{tabular}}
\vspace{-0.15in}
\end{table}

\begin{table*}[t]
\centering
\caption{We ablate each component of our FedRand and measure its performance (BLEU, ROUGE, and CIDEr) on MSCOCO dataset and robustness (MaxRény-10\%) against the membership inference attack.}
\label{tab:ablation}
\resizebox{0.9\textwidth}{!}{\begin{tabular}{lcccccccc}
\toprule
& \multicolumn{6}{c}{\textbf{MSCOCO} ($\uparrow$)} & \multicolumn{2}{c}{\textbf{MaxRényi-10\%} ($\downarrow$)}\\
\cmidrule(lr){2-7} \cmidrule(lr){8-9}
Component & BLEU-1 & BELU-2 & BLEU-3 & BELU-4 & ROUGE & CIDEr & Image & Caption\\
\midrule
$\rho=0.3$ & \underline{75.57} & 58.58 & 43.36 & 31.47 & 54.90 & 109.37 & 53.89 {\scriptsize (2.79)} & 65.53 {\scriptsize (3.07)} \\
$\rho=0.7$ & 75.23 & 58.20 & 42.97 & 31.11 & 54.86 & 108.98 & \underline{52.79} {\scriptsize (1.37)} & \underline{65.40} {\scriptsize (4.22)} \\
w/o past parameters & \textbf{76.30} & \textbf{59.29} & \textbf{44.23} & \textbf{32.42} & \textbf{55.27} & \textbf{110.83} & 58.04 {\scriptsize (5.35)} & 67.44 {\scriptsize (4.33)}\\
w/o normalization & 72.50 & 54.90 & 39.61 & 28.04 & 52.79 & 98.83 & \textbf{51.03} {\scriptsize (2.12)} & \textbf{62.21} {\scriptsize (1.71)} \\
\midrule
FedRand & 75.37 {\scriptsize (0.35)} & \underline{58.66} {\scriptsize (0.38)} & \underline{43.63} {\scriptsize (0.23)} & \underline{31.89} {\scriptsize (0.25)} & \underline{55.15} {\scriptsize (0.19)} & \underline{110.27} {\scriptsize (0.54)} & 53.84 {\scriptsize (2.50)} & 66.61 {\scriptsize (3.22)}\\
\bottomrule
\end{tabular}
}
\vspace{-0.1in}
\end{table*}

In contrast, FedPer and FedPara exhibit significantly lower performance on both the server and client sides compared to FedAvg and FedRand across the  ScienceQA, and MSCOCO datasets. This underperformance is attributed to their client-specific private parameters. Since these parameters are never aggregated, knowledge transfer between clients is limited, leading to overfitting on small client datasets and a degradation in generalization performance. On the other hand, our method, FedRand, stochastically shares a random subset of client parameters at each round, encouraging knowledge transfer between clients. This mitigates the overfitting issue and improves generalization.

\vspace{-0.1in}
\paragraph{OOD generalization.}Furthermore, we evaluate the models trained on the MSCOCO dataset using the NoCaps dataset to measure out-of-distribution (OOD) generalization performance. As shown in~\Cref{tab:vl-ood}, we observe similar trends to those in the previous experiments. FedRand achieves performance comparable to FedAvg, while FedPer and FedPara significantly degrade in performance compared to both FedAvg and FedRand. These results once again highlight the effectiveness of our method, FedRand.



\vspace{-0.1in}
\paragraph{Membership inference attack (MIA).} We perform a membership inference attack on the models trained on the MSCOCO dataset. Following~\citet{li2024membership}, we use MaxRényi-K\%, described in~\Cref{eq:max-renyi}, as a score for binary classification to distinguish member data instances in the MSCOCO dataset from non-member ones in the NoCaps dataset, and report the AUROC score in~\Cref{tab:mia-server}. A sample of 300 is drawn from each population for member and non-member data, consisting of 600 images in total. Notably, the non-member data primarily consists of object images that rarely appear in MSCOCO.

We consider two plausible scenarios: \textbf{(a)} the server attempts a MIA using the aggregated model (denoted as `\emph{server}' in the table), and \textbf{(b)} the server maliciously reconstructs the client model and performs MIA (denoted as `\emph{client}' in the table). In the case of FedAvg, the server can exactly reconstruct client models using the full client LoRA parameters transmitted to it.  However, in our FedRand, since only a subset of parameters is sent to the server per round, the timing at which a client sends the other set of parameters varies across clients. Thus, we first intercept one part of LoRA weights from each client in the final round. Then we obtain the rest of the LoRA weights at the second-to-last round in which each corresponding client participates. For FedPer and FerPara, the client model cannot be fully reconstructed under any circumstance; therefore, we report only the `\textit{server}' results for those two methods. 

\begin{table}[t]
\centering
\caption{Membership inference attack to distinguish the training dataset MSCOCO from the NoCaps dataset using Rényi Entropy Max\_0\% and Max\_10\%. \textbf{Lower} scores indicate \textbf{better robustness} against the membership inference attack. Statistics are presented in percentage. }
\label{tab:mia-server}
\resizebox{0.48\textwidth}{!}{\begin{tabular}{lcccc}
 \toprule
 & \multicolumn{2}{c}{\textbf{MaxRényi-0\%} ($\downarrow$)} & \multicolumn{2}{c}{\textbf{MaxRényi-10\%} ($\downarrow)$}  \\
 \cmidrule(lr){2-3} \cmidrule(lr){4-5} 
 & image &  caption &  image &  caption  \\
\midrule
FedAvg (server)  & 49.96 {\scriptsize (3.11)} & 70.22 {\scriptsize (2.56)} & \underline{54.57} {\scriptsize (4.07)} & 70.22 {\scriptsize (2.56)}  \\
FedAvg (client) & 51.68 {\scriptsize (4.17)} & 70.68 {\scriptsize (3.82)} & 54.71 {\scriptsize (4.11)} & 70.69 {\scriptsize (3.80)}\\
FedPer (2 layers) & 50.73 {\scriptsize (4.36)} & 70.01 {\scriptsize (3.87)} & 56.76 {\scriptsize (1.51)} & 70.03 {\scriptsize (3.84)} \\
FedPer (4 layers) &  51.77 {\scriptsize (3.40)} & 69.71 {\scriptsize (4.14)} & 57.74 {\scriptsize (2.25)} & 69.73 {\scriptsize (4.16)} \\
FedPara & 53.48 {\scriptsize (1.77)} & 69.67 {\scriptsize (2.97)} & 57.07 {\scriptsize (2.93)} & 69.63 {\scriptsize (2.99)}  \\
\midrule
\textbf{FedRand (server)} & \underline{48.90} {\scriptsize (4.75)} & \textbf{67.02} {\scriptsize (3.74)} & \textbf{53.84} {\scriptsize (2.50)} & \textbf{66.61} {\scriptsize (3.22)} \\
\textbf{FedRand (client)} & \textbf{47.83} {\scriptsize (3.56)} & \underline{68.51} {\scriptsize (3.69)} & 54.99 {\scriptsize (4.22)} & \underline{68.51} {\scriptsize (3.69)} \\
\bottomrule
\end{tabular}
}
\end{table}

As shown in \Cref{tab:mia-server}, FedRand demonstrates stronger resistance to MIA compared to the other baseline methods. This is due to the fact that clients send only a subset of parameters to the server, which helps prevent the exposure of their full client parameters. Both FedAvg and FedRand show that reconstructed client models are more vulnerable than server models, with this trend being more pronounced in FedAvg, as it can fully reconstruct client models at the end of any round. FedPer and FedPara are expected to be effective against MIA since they do not share client-specific private parameters at all; however, they show worse robustness than FedRand. This may be attributed to the fact that their private parameters are never shared across clients, limiting knowledge transfer. As a result, the shared global parameters must compensate by fitting each client's dataset more closely, making them more prone to overfitting and leading to more severe memorization.

\paragraph{Ablation studies.} We conduct a comprehensive ablation study on each component of our method to evaluate its effectiveness. First, we vary the probability $\rho$ of selecting the LoRA weight matrix $A$, setting it to $\rho=0.3$ and $\rho=0.7$. Additionally, we ablate the normalization factors $\alpha$ and $\beta$, as defined in~\Cref{eq:normalization}, referring to this case as ``w/o normalization.'' Lastly, instead of using the client-specific private parameters in lines 10 and 13 of~\Cref{algo-client}, we initialize with the full LoRA weights from the server and send either the updated $A$ or $B$ back to the server, depending on the variable $a_k$, referring to this case as ``w/o past parameters''.  

As shown in~\Cref{tab:ablation}, selecting the LoRA weight matrix 
$A$ either more or less frequently than $B$ degrades the performance of image captioning on MSCOCO while slightly improving robustness against MIA. Similarly, removing normalization significantly degrades BLEU, ROUGE, and CIDEr scores, while making the model more robust to MIA due to underfitting. In contrast, initializing all the client parameters with the LoRA weights of the server without using the client's past parameters significantly boosts the performance on the MSCOCO dataset but drastically sacrificing robustness against the MIA. These experimental results support the choice of hyperparameters $\rho=0.5$ and our algorithm design. 

\begin{figure}[t]
    \includegraphics[width=0.48\textwidth]{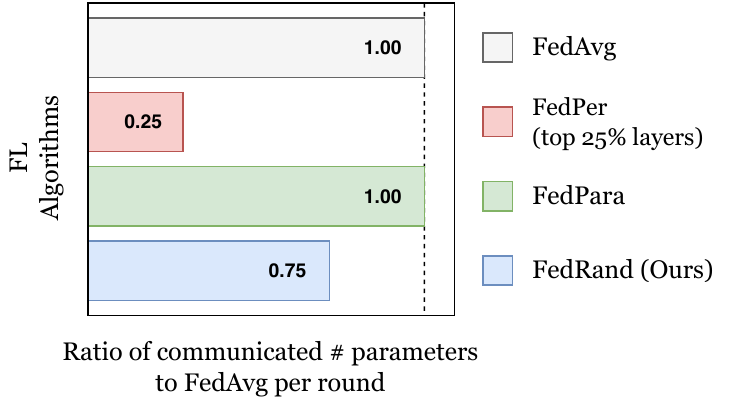}
    \caption{The ratio of number of communicated LoRA parameters, compared to FedAvg per round under LoRA configuration.}
    \label{fig:commu-cost}
\end{figure}

\paragraph{Communication cost.} \Cref{fig:commu-cost} illustrates the communication cost between a server and clients required for each method. Although FedPer reduces the the cost to 25\% by sharing only the  upper layers, it significantly underperforms compared to FedAvg as shown in previous experiments. In the case of our proposed FedRand, receives the same number of parameters received from the server as FedAvg, but only sends half of them are back to the server, reducing the communication cost by approximately 25\% per round, while retaining accuracy similar to FedAvg.

\section{Conclusion}
In this work, we proposed the FedRand framework to mitigate the vulnerability of vision-language models (VLMs) fine-tuned with federated learning to membership inference attacks. Instead of communicating the full LoRA weights of VLMs between the server and clients --- which an attacker could intercept to perform membership inference attacks --- each client randomly selected a subset of LoRA weights from the server and initialized the remaining LoRA weights using its private parameters from the previous round. After updating both sets of parameters, only the non-private parameters were sent back to the server for aggregation, reducing the risk of disclosing the full parameters of the client model. We extensively validated that our proposed FedRand achieved performance comparable to FedAvg, which communicated full LoRA weights between the server and clients, while demonstrating improved robustness against membership inference attacks compared to other relevant baselines. Additionally, our method reduced communication costs between the server and clients by transmitting only a subset of the client model parameters to the server. As future work, we suggested randomly selecting sub-layers of clients for training or quantizing client parameters sent to the server to further enhance the security of client model parameters. 

\section*{Impact Statements}
This paper presents a framework, FedRand, aimed at improving privacy in Federated Learning (FL), particularly when training vision-language models (VLMs). Our work contributes to advancing the field of privacy-preserving machine learning by mitigating the risks of membership inference attacks without significantly compromising model performance. By enhancing data privacy in FL, our approach can benefit various real-world applications, including healthcare, finance, and other domains where sensitive data is distributed across multiple entities. FedRand reduces the exposure of client-side model parameters, thereby strengthening privacy guarantees for users participating in federated training. However, as with any privacy-preserving method, FedRand does not eliminate all risks. Adversarial attackers may still attempt more sophisticated attacks beyond membership inference, and further research is needed to address emerging privacy threats. Additionally, while our method enhances privacy, it does not directly address fairness or bias in FL, which remain important considerations for real-world deployment. Overall, this work aligns with the broader goal of developing privacy-preserving AI systems and does not introduce any foreseeable ethical concerns or negative societal impacts.
\bibliography{references}
\bibliographystyle{icml2025}

\newpage
\appendix
\onecolumn
\section{Experimental Details}

\subsection{Dataset}
\label{sec:appendix-a-1}
\begin{itemize}
    \item \textbf{ScienceQA}~\citep{scienceqa} is a  multiple choice visual question answering dataset derived from elementary and high school science curricula, covering three subjects: natural science, language science, and social science. We focus exclusively on the 10,327 questions that include accompanying images, representing 48.7\% of the entire dataset.

    \item \textbf{MSCOCO}~\citep{mscoco} contains over 330K images with dense annotations for image recognition, segmentation and captioning tasks. Among the 83K instances specifically created for captioning, 50K images are sampled for training and 5K images each for validation and testing. 

    \item \textbf{NoCaps}~\citep{nocaps} is designed to evaluate the ability of image captioning models to describe objects not present in the MSCOCO dataset. 45K validation sets, each with 10 captions, are used to assess OOD generalization.
\end{itemize}

\subsection{Prompt Template}
We present a prompt template for each dataset. Note that the presence of contextual information in ScienceQA depends on the question.

\resizebox{0.95\textwidth}{!}{\begin{prompt}{ScienceQA}
        \footnotesize
        \texttt{Based on the image, respond to the question with a given options.} \\
        \texttt{\textbf{USER}: \{image\}\textbackslash n Context: \{context\}. Options: \{options\}. Answer:}  \\
        \texttt{\textbf{ASSISTANT}: ...} \\
\end{prompt}}
\resizebox{0.95\textwidth}{!}{\begin{prompt}{MSCOCO \& NoCaps}
        \footnotesize
        \texttt{Briefly describe given image.} \\
        \texttt{\textbf{USER}: \{image\}\textbackslash n A short image description:}  \\
        \texttt{\textbf{ASSISTANT}: ...} \\
\end{prompt}}
\label{sec:appendix-a-3}

\subsection{Communication process of FedPer and FedPara}
\label{sec:appendix-a-4}
In the original FedPer framework, the classifier and top $N$ basic blocks of a ResNet~\citep{resnet} model are designated as personalization layers. To adapt this approach for LoRA settings, we instead share the LoRA parameters of the top 2 or 4 transformer~\citep{transformer} layers with the server.

Similarly, the FedPara method originally parameterize weight of base models with Hadamard product between two sets of low rank matrices.  To extend this idea to transformer architecture LLMs with LoRA,  we introduce an additional pair of LoRA A and B matrices per layer, ensuring the additional LoRA weight matrices remain private on the client side.

\clearpage
\section{Membership Inference Attack Example}
\begin{figure}[t]
\centering
    \includegraphics[width=0.9\textwidth]{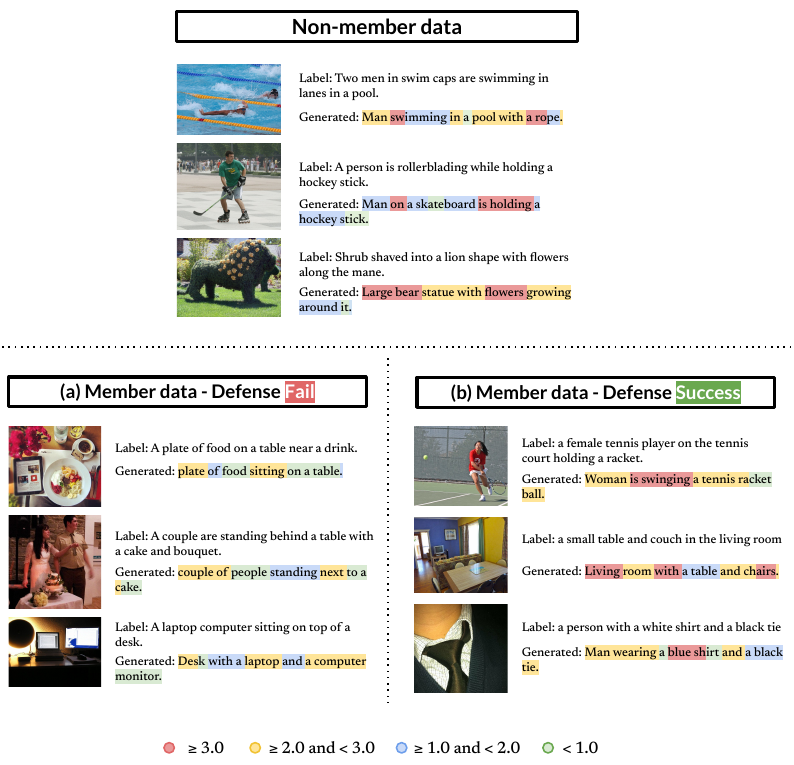}
    \vspace{-0.1in}
    \caption{An example of token-wise Rényi entropy measurement for member (MSCOCO) and non-member (NoCaps) data. The higher the entropy is, the more robust to MIA.}
    \label{fig:mia-example}
\vspace{-0.1in}
\end{figure}

We show two sets of membership inference attack (MIA) examples in~\Cref{fig:mia-example}, where color denotes token-wise Rényi entropy with FedRand. On the left (a), the model is confident in next-token prediction for member data (MSCOCO), indicating a failed defense against MIA. On the right (b), the model is highly uncertain for both member and non-member data (NoCaps), leading to a successful defense against MIA.
\label{sec:appendix-b}

\end{document}